\documentclass[conference]{IEEEtran}
\IEEEoverridecommandlockouts
\usepackage{cite}
\usepackage[ruled,linesnumbered]{algorithm2e}
\usepackage{amsmath,amssymb,amsfonts}
\usepackage{algorithmic}
\usepackage{graphicx}
\usepackage{textcomp}
\usepackage{xcolor}
\usepackage{url}
\def\BibTeX{{\rm B\kern-.05em{\sc i\kern-.025em b}\kern-.08em
    T\kern-.1667em\lower.7ex\hbox{E}\kern-.125emX}}
\begin{document}

\title{Joint Progressive Knowledge Distillation and Unsupervised Domain Adaptation}

\author{\IEEEauthorblockN{Le Thanh Nguyen-Meidine, Eric Granger, Madhu Kiran, Jose Dolz}
\IEEEauthorblockA{\textit{Laboratoire d'imagerie, de vision et d'intelligence artificielle (LIVIA)} \\
\textit{Dept. of Systems Engineering, \'Ecole de technologie sup\'erieure}\\
Montreal, Canada \\
\{le-thanh.nguyen-meidine.1, madhu.kiran.1\}@ens.etsmtl.ca \\
\{eric.granger, jose.dolz\}@etsmt.ca}
\and
\IEEEauthorblockN{Louis-Antoine Blais-Morin}
\IEEEauthorblockA{\textit{Genetec Inc.} \\
Montreal, Canada \\
lablaismorin@genetec.com}
}

\maketitle

\begin{abstract}
Currently, the divergence in distributions of design and operational data, and large computational complexity are limiting factors in the adoption of CNNs in real-world applications. For instance, person re-identification systems typically rely on a distributed set of cameras, where each camera has different capture conditions. This can translate to a considerable shift between source (e.g. lab setting) and target (e.g. operational camera) domains. Given the cost of annotating image data captured for fine-tuning in each target domain, unsupervised domain adaptation (UDA) has become a popular approach to adapt CNNs. Moreover, state-of-the-art deep learning models that provide a high level of accuracy often rely on architectures that are too complex for real-time applications. Although several compression and UDA approaches have recently been proposed to overcome these limitations, they do not allow optimizing a CNN to simultaneously address both. 
In this paper, we propose an unexplored direction -- the joint optimization of CNNs to provide a compressed model that is adapted to perform well for a given target domain. In particular, the proposed approach performs unsupervised knowledge distillation (KD) from a complex teacher model to a compact student model, by leveraging both source and target data. It also improves upon existing UDA techniques by progressively teaching the student about domain-invariant features, instead of directly adapting a compact model on target domain data. 
Our method is compared against state-of-the-art compression and UDA techniques, using two popular classification datasets for UDA -- Office31 and ImageClef-DA. In both datasets, results indicate that our method can achieve the highest level of accuracy while requiring a comparable or lower time complexity. 
\end{abstract} 

%

\begin{IEEEkeywords}
Deep Learning, Convolutional Neural Networks, Domain Adaptation, Knowledge Distillation, Visual Recognition.
\end{IEEEkeywords}

\section{Introduction}

Deep learning (DL) models, and in particular convolutional neural networks (CNNs) can achieve state-of-the-art performance in a wide range of visual recognition applications, such as classification, object detection, and semantic segmentation \cite{yuan2019objectcontextual, Mask_RCNN, alex2019large}. In practice, a main drawback associated with these models is their scalability and computationally complexity, which poses a challenge for many real-time applications, as found in video analytics and surveillance \cite{IPTA_CNN}. 

Currently, the compact DL models that can provide fast inference generally lack the high accuracy of deeper, more complex models. One alternative to overcome this issue is to compress complex high accuracy models into smaller or simpler models while preserving the same level of accuracy. Several approaches have recently been proposed to accelerate and compress CNNs, include  quantization,  low-rank approximation, knowledge distillation, compact network design and network pruning. For instance, network compression methods for channel pruning \cite{Lottery, HugoSurvey,  Molchanov_2019_CVPR, ICPR_Pruning, NetworkPruningviaTransformableArchitectureSearch, liu2018rethinking} and knowledge distillation (KD) \cite{HIntonKD, RelationalKD, Overhaul, TeachingAssistantKD}) have become very popular due to the exponential increase in the complexity of architectures, which may have millions of parameters.

Another limitation is the poor generalization of CNNs across domains, particularly when there is a  considerable domain shift between source and target data distributions. This is the case for instance of applications like video-surveillance over a distributed network of camera, where variations in camera viewpoint and capture conditions (e.g., illumination, occlusion and background) introduce a shift between data from source and target domains. The accuracy of CNNs degrades when there is a considerable divergence between the data capture conditions in the model development and operational environments. To alleviate this problem, domain adaptation techniques are commonly proposed, either in a supervised or unsupervised setting \cite{DA_Survey, wangsurvey}. For applications in video-surveillance, it is, however, costly to collect and annotate videos from each camera viewpoint and capture condition to fine-tune a CNN.  

In this paper, we focus on DL models for unsupervised domain adaptation (UDA) to allow adapting CNN embeddings based on unlabeled data. 
The main body of literature on UDA techniques focuses on learning domain invariant features by using adversarial loss \cite{GRL, ADDA} to encourage domain confusion, or by minimizing a distance or discrepancy between two the data distributions \cite{MMD_ICLR}, or both \cite{WD_DA_GAN}. Another popular paradigm is to learn a mapping between source and target images such that images captured in different domains have a similar appearance. This reconstruction-based approach mimics standard supervised learning \cite{DomainMapping1, DomainMapping2}. 

While compression and UDA techniques can respectively provide CNNs with a high level of performance and efficiency, research that explores the benefits of joint model compression and UDA remains scarce. For example, recent research has combined KD and UDA to improve the performance in a UDA context \cite{TeachingToAdapt, orbesarteaga2019knowledge}. Despite the improvement shown in terms of the DA task, they neglect the reduction of model complexity. Another important problem when employing KD is the potentially large gap between the capacity of teacher and student models to learn complex mappings, which can degrade student performance. Distilling knowledge directly into the student model from a trained teacher model may present some challenges. For example, in the early stages of training, the decision boundaries of the student and teacher modes may differ considerably. To bridge this gap some researchers have proposed to either add a teaching assistant\cite{TeachingAssistantKD} or distilling knowledge at several intermediate layers of a CNN \cite{Overhaul}. 

In a standard training scenario, the CNN model would be adapted to a new target after having been compressed, although this can reduce its capacity to generalize on target data since over-parametrization is often important for generalization \cite{NIPS2019_8847}. This can be overcome by learning from a teacher instead of learning directly through the UDA loss. In contrast, another scenario involves adapting a complex CNN model to the target domain, and then compressing it, although unsupervised KD remains a challenge since ground truth data is needed to assure that the student model to remain consistent with the dataset. To address the aforementioned limitations, we argue that by adapting the teacher to the target domain while the student is being trained, the student can adapt progressively to the domain instead of learning directly from a previously targeted domain. The student would thereby learn the steps needed to adapt itself to the target domain, instead of learning directly from a model adapted to the target, i.e. adapted teacher. By jointly exploiting unsupervised KD and DA, it is possible to overcome the lack of ground truth data for KD, as needed to train the student model that is consistent with source features.  

Unlike recent work in literature, this paper provides the first attempt to simultaneously address two key problems with CNNs -- domain shift and model complexity -- through joint progressive KD and UDA. Particularly, the proposed approach learns a compact model with a feature embedding that can provide a high level of accuracy in the target domain. It leverages progressive KD to adapt the student model in a step-by-step manner, using knowledge from the teacher to learn domain invariant features of the source and target domains. In order to ensure the validity of student model w.r.t the UDA loss and the target domain, we introduce a consistency loss that ensures consistency on the student model by learning source domain features from the teacher. We validate our approach with different training scenarios of KD and UDA: (1) UDA then KD, (2) KD then UDA and (3) UDA directly on compact model.
Empirical evaluations show that our joint progressive KD and UDA approach facilitates domain adaptation and compression of deep CNNs, and can outperform representative state-of-the-art approaches on the widely used Office31 and ImageClef-DA benchmark datasets. In particular, our approach is general and model-agnostic, and can be combined with different UDA and KD techniques to improve performance. 


\section{Related Work}

\subsection{Compression techniques:}

Approaches for compressing CNNs can be mainly categorized in: (1) pruning techniques\cite{Lottery , Molchanov_2019_CVPR, NetworkPruningviaTransformableArchitectureSearch, liu2018rethinking, PruningFPGM}, (2) quantization\cite{Quanti1, Quanti2, Quanti3}, (3) decomposition\cite{LRA_Jader, Coordinating}, and (4) KD\cite{HIntonKD, RelationalKD, Overhaul, TeachingAssistantKD}. Pruning techniques focus on removing non-useful weights or filters in order to reduce the computational complexity. 
Quantization techniques focus on reducing the representation of weights into lower precision since, for example, 8-bit integer precision provides much faster computation than the floating point computation. Decomposition techniques provide faster computation by decomposing tensor in lower rank approximation as vectors products. Lastly, KD techniques transfer knowledge from a teacher (usually a large model) to a student (a smaller model). Since we will employ this technique to reduce the complexity of the model, we will focus on KD onward. For a comprehensive survey on compression techniques for CNNs, we refer the reader to \cite{cheng2017survey}. 

There exist several ways of distilling knowledge from a teacher to a student. A well-known technique is to employ the teacher output as the soft label for the student \cite{HIntonKD}. In this work \cite{HIntonKD}, the temperature value was employed to generate softer versions of the teacher outputs. Another popular solution is to minimize the features differences at intermediate layers between the teacher and student network in order to maximize the information transfer between the teacher and the model \cite{Overhaul}. Feature similarity can be enforced by minimizing a partial L2 distance, which is equivalent to L2 norm except that if the value of the student is smaller than the teacher and both are negative then the result is zero, between student and teacher after using a Margin ReLU (use of a margin $m$ instead of $0$) which can take in negative values of a feature map. To solve the gap issue between a converged teacher model and a student some other researchers have proposed the integration of a teaching assistant \cite{TeachingAssistantKD}. In this work, an intermediate model with a lesser gap is chosen, this model is then used as a teacher to a smaller model, by progressively reducing the gap with an intermediate teaching assistant, thus, limit the performance degradation of the student model.  

\subsection{Unsupervised domain adaptation:}



Unsupervised Domain Adaptation (UDA) techniques try to adapt models when a domain shift between source and target dataset exists, where only the source data is labeled. Current main UDA techniques \cite{DA_Survey} include: finding domain-invariant features\cite{MMD_ICLR}, domain mapping\cite{DomainMapping_Pixel_Level, ConGAN_DA_MAP}, ensemble learning\cite{Tri_net_DA_Ensemble}, statistic normalization\cite{DA_Normalization_statistics} and target discriminate methods\cite{Wei2018GenerativeAG}. The first category learns domain invariant features either by domain confusion\cite{GRL, ADDA} or minimizing a distance between distribution \cite{MMD_ICLR}. Domain confusion can be achieved by employing a domain classifier (or discriminator) \cite{GRL, ADDA}. While \cite{GRL} employs a gradient reversal layer in order to maximize the domain classification loss, the work in \cite{ADDA} uses an adversarial loss on the discriminator. Domain mapping focuses on finding a mapping either from the source domain to the target domain or vice-versa \cite{DomainMapping_Pixel_Level, ConGAN_DA_MAP}. Currently, most of the domain mapping based techniques rely on generative adversarial networks (GAN). Hoffman et \textit{al.} \cite{DomainMapping_Pixel_Level} propose to use a pair of discriminator and generator in order to map a source image into the target domain distribution. This mapping is learned at the same time as a task-specific loss (i.e. classification loss) on both transformed image and non transformed-images and the overall optimization is done alternatively between generator and discriminator-task. The paper in \cite{DomainMapping_CADA} goes further by integrating adaptation at feature-level. Ensemble methods use either multiple models or the same model at different times (typically referred to as self-ensembling) in order to produce more reliable pseudo-labels on unlabelled data \cite{Tri_net_DA_Ensemble}. Others methods like statistic normalization assume that the task knowledge is learned and the only adaptation needs to be done is on the batch norm statistics\cite{DA_Normalization_statistics}. Last, target discriminate methods work with the assumption that data points are distributed in separate clusters and the decision boundary lies in lower density regions. Thus, these methods work by trying to push the decision boundary to lower density regions by adding adversarial losses \cite{Wei2018GenerativeAG}. 


 

\subsection{Joint unsupervised domain adaptation and knowledge distillation:}

Even though jointly exploiting UDA and KD for compression and domain shift problems remains unexplored, there have been few attempts to combine these two techniques in the context of domain adaptation \cite{KnowledgeAdaptation,SSDA_KD_MRI}. In \cite{KnowledgeAdaptation}, the authors propose to use multiple teachers for teaching a single student in order to increase the performance of their model in the context of sentiment analysis. The work in \cite{SSDA_KD_MRI} proposes to combine KD and DA for the task of white matter hyperintensities segmentation in magnetic resonance imaging by training the teacher model on the source and trying to minimize the cross-entropy loss between the probability maps of the teacher and the student on the target dataset. Nevertheless, these approaches do not address the problem of reducing model complexity. To the best of our knowledge, there exist only one approach that combines compression and UDA techniques, referred to as TCP \cite{TCP}. Interaction between the two techniques is done in several steps. First, the model is trained to be adapted to the target domain by minimizing the domain divergence using the maximum mean discrepancy (MMD) \cite{MMD_ICLR}. Then, the least important filters are pruned by using a gradient based criterion and the model is continuously refined on the domain adaptation loss. An important limitation of this technique is the need to have an already domain adapted model in order to start the pruning, whereas our technique can directly start from a non-adapted model. 


\section{Proposed Method}





\begin{figure*}[h!]
    \centering
    \includegraphics[width=1.0\textwidth]{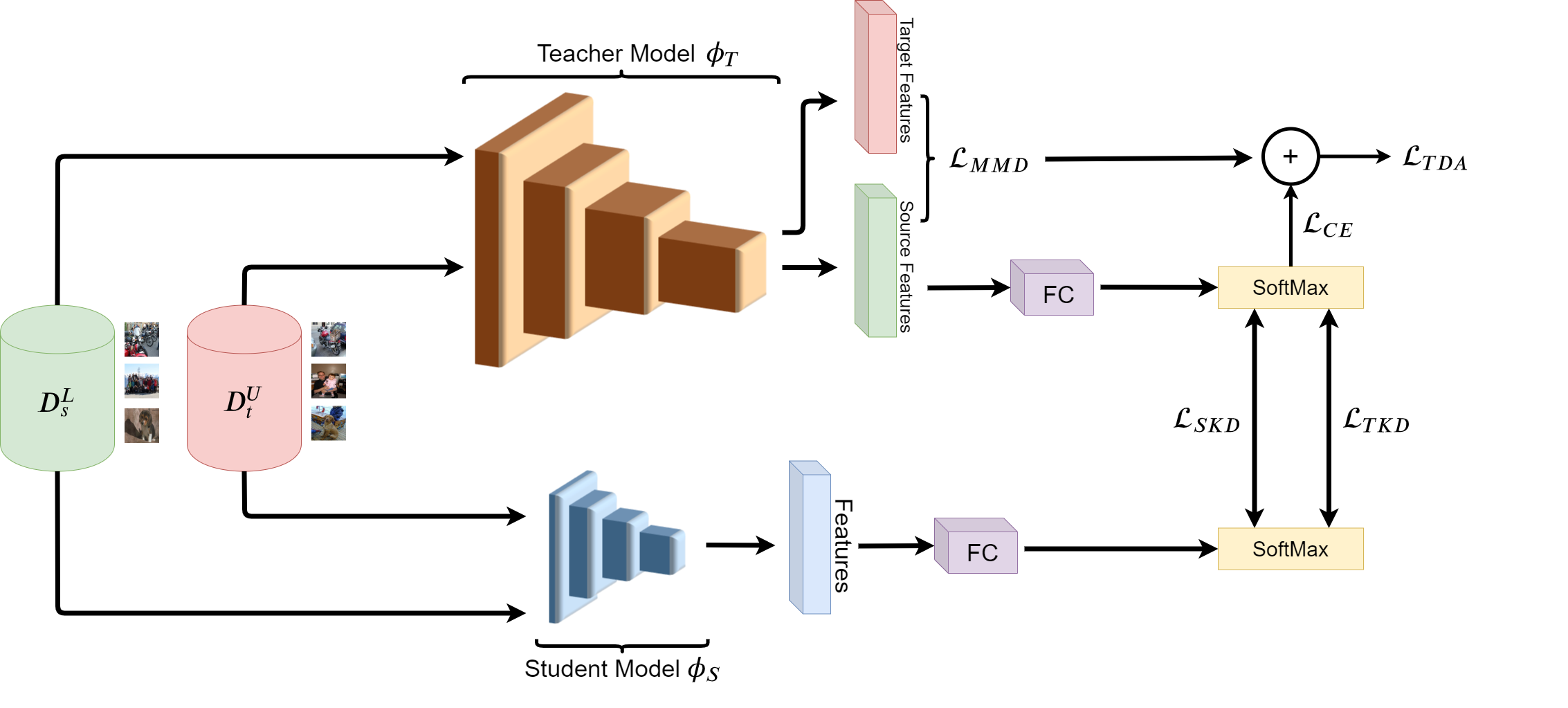}
    \caption{Illustration of the proposed learning technique for progressive KD with UDA.}
    \label{fig:KD-UDA}
\end{figure*}

The main pipeline of the proposed method is depicted in Figure \ref{fig:KD-UDA}. Our method performs domain adaptation of a teacher model by learning domain invariant features between the source and target domains. Meanwhile, it progressively distills its knowledge to a student model on both source and target features. As shown in Figure \ref{fig:KD-UDA}, DA is performed on the features of the teacher network, while the KD from teacher to student is performed on the result of a temperature-based soft-max on the logits (output of a fully connected layer). Additional details of the proposed UDA and KD techniques are described in the following subsections.


\subsection{Unsupervised domain adaptation:}

We start by defining the UDA loss for teacher model, which is based on MMD \cite{MMD_ICLR, TCP}\footnote{Note that it can be generalized to most other UDA techniques.}:

\begin{equation}\label{eq:teacher_mmd}
	\begin{aligned}
	\mathcal{L}_{MMD} = || \frac{1}{N_s} \sum_{x_i \in D^{L}_s} \phi_T(x_i) - \frac{1}{N_t} \sum_{x_j \in D^{U}_t} \phi_T(x_j)||_\mathcal{H}^2
	\end{aligned}
\end{equation}

With $D^{L}_s$ the labeled source domain dataset which contains $N_s$ samples and labels, $D^{U}_t$ the unlabeled target domain dataset of $N_t$ data samples, $\phi_T$ the teacher feature extractor function that maps an input to a feature map, $\mathcal{H}$ the Reproducing Kernel Hilbert Space(RKHS) with gaussian kernel. As in \cite{MMD_ICLR, TCP}, we incorporate this a supervised loss on the source domain in order to have the final UDA loss for the teacher:

\begin{equation}\label{eq:teacher_da}
	\begin{aligned}
	\mathcal{L}_{TDA} = \mathcal{L}_{MMD} + \gamma \mathcal{L}_{CE}(T(D^{L}_s, 1), y_s)
	\end{aligned}
\end{equation}

 $\mathcal{L}_{CE}$ the supervised cross-entropy loss of the teacher model on the source domain, $\gamma$ a trade-off hyper-parameter that follows the same variations as \cite{TCP} and $T$ the function that maps an input to the output of the teacher network with a soft-max of temperature 1 (i.e. the regular soft-max).

 \subsection{Knowledge distillation for domain knowledge transfer:}
 
 The next step is to transfer the target domain knowledge from the teacher to the student, we use a modified version of the KD loss from the work of Hinton:

\begin{equation}\label{eq:target_kd}
	\begin{aligned}
	\mathcal{L}_{TKD} = \mathcal{L}_{distill}(S(D^{U}_t, \tau), T(D^{U}_t, \tau))
	\end{aligned}
\end{equation}

In this equation, $S$ and $T$ represent respectively the output of student network and teacher network with a soft-max based on a temperature $\tau$ in order to soften the output and $L_{distill}$ is a KL divergence loss in our case but can be replaced with a mean squared loss or cross-entropy. This loss differs from the original paper\cite{HIntonKD} because, we had to remove the cross-entropy loss between the student model output and the ground truth since we are working on UDA. Trivially, this should be enough for joint KD and UDA since we only want to have target domain knowledge, in order to ensure the consistency of the model w.r.t to a common representation, we proposed to add a consistency loss to ensure that the student model can learn a better common representation from source and target domains by distilling on the source data.

\begin{equation}\label{eq:source_kd}
	\begin{aligned}
	\mathcal{L}_{SKD} = \mathcal{L}_{distill}(S(D^{L}_s, \tau), T(D^{L}_s, \tau)) + \alpha  \mathcal{L}_{CE}(S(D^{L}_s, 1), y_s)
	\end{aligned}
\end{equation}

Eq. \ref{eq:source_kd} is the student KD loss, with hyper-parameter $\alpha$ to balance between the KD and the cross entropy loss of the output of the student model and the ground truth on the source domain. The Figure \ref{fig:KD-UDA} illustrates all these losses and also the proposed techniques. The final loss of our models, is then:

\begin{equation}\label{eq:total_loss}
	\begin{aligned}
	\mathcal{L} = (1 - \beta)\mathcal{L}_{TDA} + \beta(\mathcal{L}_{TKD} + \mathcal{L}_{SKD})
	\end{aligned}
\end{equation}

We added the $\beta$ hyper-parameter in order to balance out the importance between UDA and KD. Since we are performing jointly KD and DA, in the beginning, the teacher would still be learning from the DA. This means that there is not much to be learned for the student model, besides the source representation which can be learned from the KD loss. In light of this, we propose to start by giving more importance to UDA in the beginning and gradually transfer the importance to KD basing $\beta$ on an exponential growth function between $[b, f]$, with b the starting value of $\beta$ and $f$ the end value. In order to define as exponential growth, we need to calculate a growth rate based on $b$ and $f$:

\begin{equation}\label{eq:growth_rate}
	\begin{aligned}
	g = \frac{\log(\frac{f}{b})}{epochs}
	\end{aligned}
\end{equation}

With $epochs$ the number of epochs and $g$ the growth rate. Once we have the growth rate, $\beta$ at epoch $t$ can be found as:
\begin{equation}\label{eq:update_beta}
	\begin{aligned}
	\beta_t = b * e^{gt}
	\end{aligned}
\end{equation}

\begin{algorithm}[h]
\SetAlgoLined
\caption{KD-UDA}
\label{KD-UDA-Algo}
\SetKwInOut{Input}{input}
\SetKwInOut{Output}{output}
\SetKwInOut{Parameter}{parameter}
\Input{A teacher model $M_T$, a student model $M_S$, a source dataset $D^{Sup}_s$, a target dataset $D^{U}_t$}
\Output{A target adapted student model}

\For {$epoch\gets{1}$ \KwTo {epochs}} {
    \For {$data_s$ in $D^{L}_s$ and $data_t$ $D^{U}_t$} {
        Obtain the feature map $F_s$ of $M_t$ on $data_s$ and $F_T$ on $data_t$ using $\phi_T$\\
        Optimize the teacher model $M_T$ with $(1-\beta)\mathcal{L}_{TDA}$ using $F_s$ and $F_t$ \\
        Obtain the logits $o^{T}_s$ of $M_T$ on $data_s$ and $o^{T}_t$ of $M_T$ on $data_t$\\
        Obtain the logits $o^{S}_s$ of $M_S$ on $data_s$ and $o^{S}_t$ of $M_S$ on $data_t$\\
        Applying soft-max of temperature $\tau$ on the logits and optimize $\beta(\mathcal{L}_{TKD} + \mathcal{L}_{SKD})$\\
        Update $\beta$ following Eq.\ref{eq:update_beta}
    }
    Evaluate the model\\
}

\end{algorithm}
\vspace{-3.8pt}
The overall algorithm is described in Algorithm \ref{KD-UDA-Algo} using an alternate optimization scheme of our algorithm. The details of this implementation can be found in Section IV(C).  


\section{Experimental Methodology}

In this section, we detail the experimental methodology employed to validate the proposed method. First, we describe the datasets and baselines methods. Then, we provide implementation details to facilitate the reproducibility of the reported results.

\subsection{Datasets:}

\paragraph{Office31}: This dataset contains three subsets of dataset which are Webcam (W), DSLR (D) and Amazon (A) with 31 classes. These subsets contains images from amazon.com (A) or office environment with changes in lighting, poses using a DSLR camera (D) or a webcam (W). We evaluate our results based on six scenarios: $A \xrightarrow{} W$, $W \xrightarrow{} A$, $D \xrightarrow{} W$, $W \xrightarrow{} D$, $D \xrightarrow{} A$, $A \xrightarrow{} D$. 


\begin{figure}[h!]
    \centering
    \includegraphics[width=0.42\textwidth]{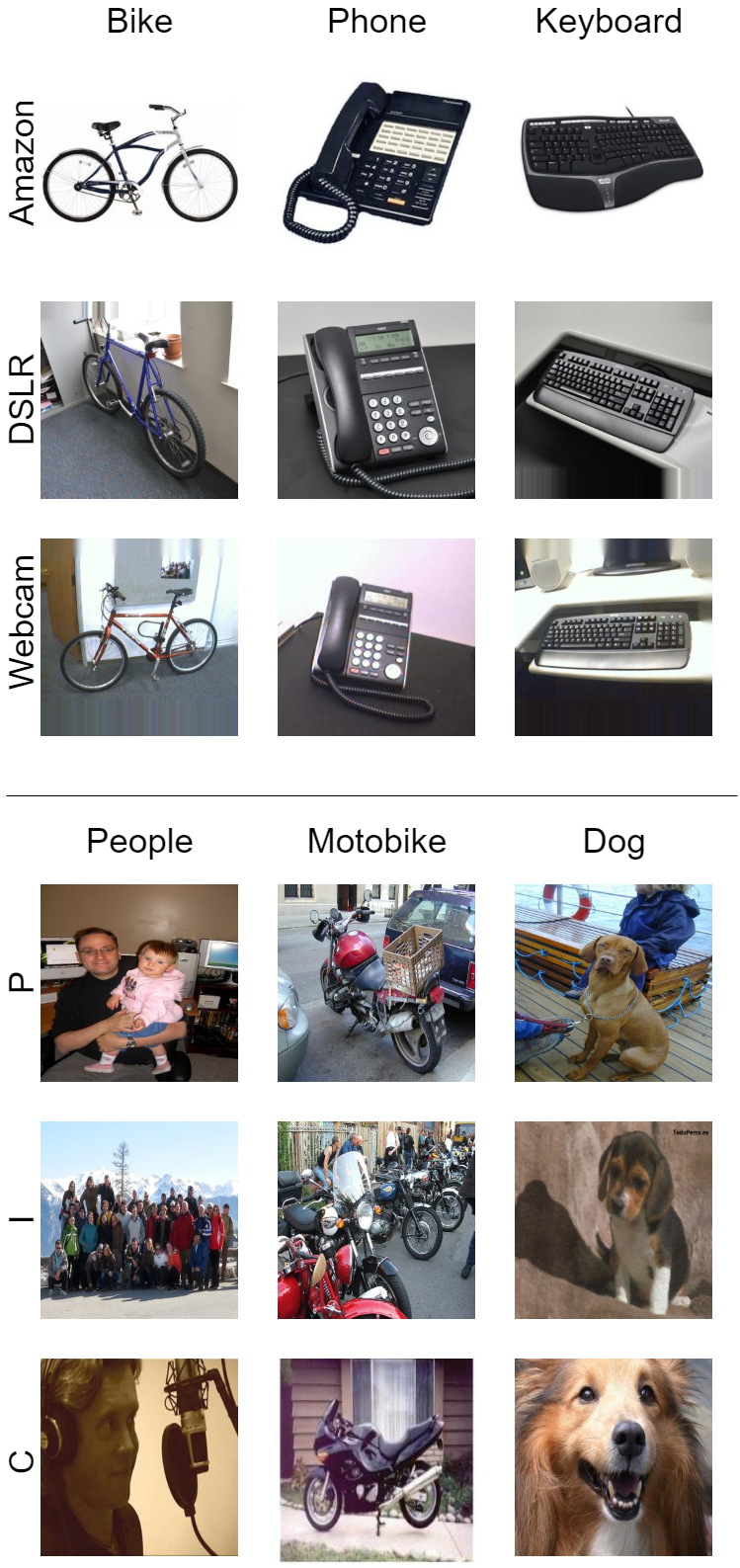}
    \caption{Examples of images selected from the Office31 (top) and ImageClef-DA (bottom) datasets.}
    \label{fig:datasets}
\end{figure}

\paragraph{ImageClef-DA}: This dataset for UDA contains four subsets which are taken from Imagenet (I), Pascal-Voc (P), Caltech (C) and Bing (B). Each of these subsets contains a total of 600 images for 12 classes. For this dataset, we compare to others techniques using six scenarios: $I \xrightarrow{} P$, $P \xrightarrow{} I$, $I \xrightarrow{} C$, $C \xrightarrow{} I$, $C \xrightarrow{} P$, $P \xrightarrow{} C$. For the evaluation of our algorithm, we chose to use a popular backbone architecture, ResNet50, with our algorithm. 


\subsection{Baselines methods:}


In order to validate our joint progressive method, we propose to evaluate and compare our method with 3 baseline scenarios: 
\begin{enumerate}
    \item UDA, and then KD,
    \item KD, and then UDA, and 
    \item UDA directly on compact model.
\end{enumerate}
With the first baseline, $UDA \xrightarrow{} KD$, we start with a model of ResNet50 for UDA, then we start KD  on this model with a modified version (Eq.\ref{eq:target_kd}) of \cite{HIntonKD} since most KD algorithms does not handle unsupervised KD. The student model used for this first baseline is similar to the one we chose for our methods, i.e. ResNet18 and ResNet34. For the second baseline $KD \xrightarrow{} UDA$, we start by training a teacher model, ResNet50, on the source dataset then we apply KD with this teacher using labeled data of the source dataset and a student of one of target model, ResNet18 or ResNet34, and we finish with UDA on this student model. As for the last baseline, we take a student model, ResNet18 and ResNet34, and directly apply UDA algorithm on it \cite{TCP, MMD_ICLR}. 

Another aspect of our method is knowledge distillation which is meant for model reduction and acceleration. In order to validate our findings, we also measure the difference in terms of FLOPs and parameters between our method and TCP. Since our models are predetermined architecture, we measure FLOPS and the number of parameters on ResNet50, ResNet34 and ResNet18, as for TCP, we report the number of FLOPS that was reported in the original paper.

\subsection{Implementation details:}


In this paper, we consider two ways of implementing the optimization process. The first is an end-to-end training by optimizing the loss of Eq.\ref{eq:total_loss} with a unique single optimizer. While this approach works and provide reasonable performance, we found that having two optimizers allows more flexibility in terms of scheduling the learning rate since there is can be a difference between optimizing domain adaptation and KD, i.e. different learning rates for each loss or we do not want to update the teacher during KD. Thus, we implement a second alternative approach where we use one optimizer for $(1 - \beta)\mathcal{L_{TDA}}$ with a different learning rate schedule than another optimizer used for $\beta(\mathcal{L_{TKD}} + L_{SKD})$. We would like to point out that it's also possible to implement the second approach with one single optimizer by handling the learning rate scheduling ourselves. For our experiments on Office31 and ImageClef-DA, we use two student models architecture, which are: ResNet34 (12\% FLOPS reduction from ResNet50) and ResNet18 (56\% FLOPS reduction from ResNet50) which are closely similar to the FLOPS reduction of TCP\cite{TCP}.

\begin{table*}[t!]
\caption{Accuracy of proposed and baseline methods on Office31 dataset when ResNet34 is the desired model.}
\label{tb:Office31_result_34}
\centering
\resizebox{\textwidth}{!}{
\begin{tabular}{|l||r|r|r|r|r|r||r|}
\hline
\textbf{Training methods}           & \textbf{A $\xrightarrow{}$ W} & \textbf{W $\xrightarrow{}$ A} & \textbf{D $\xrightarrow{}$ W} & \textbf{W $\xrightarrow{}$ D} & \textbf{D $\xrightarrow{}$ A} & \textbf{A $\xrightarrow{}$ D} & \textbf{Average}     \\ \hline \hline
Baseline 1: UDA $\xrightarrow{}$ KD  from ResNet50  & 25.4   & 7.1    & 28.5   & 50.0   & 9.7    & 30.7   & 25.2        \\ \hline
Baseline 2: KD $\xrightarrow{}$ UDA from ResNet50 & 75.7 & 61.2 & 97.8 & 99.7 & 59.6 & 81.1 & 79.1 \\ \hline
Baseline 3: UDA only on ResNet34        & 67.2   & 52.3   & 93.6   & 96.6   & 52.2   & 71.6   & 72.2       \\ \hline \hline
TCP: 12\% pruned from ResNet50 & 81.8   & 55.5   & 98.2   & 99.8   & 50     & 77.9   & 77.2        \\ \hline
Ours: ResNet34 from ResNet50       & 85.7   & 62.3   & 97.1   & \textbf{100}    & \textbf{61.8}   & 82.1   & 81.5 \\ \hline
Ours: ResNet34 from ResNet101      & \textbf{87.5}   & \textbf{62.9}   & \textbf{98.1}   & \textbf{100}    & 60.8   & \textbf{85.7}   & \textbf{82.5}        \\ \hline
\end{tabular}
}
\end{table*}

\begin{table*}[t!]
\caption{Accuracy of proposed and baseline methods on Office31 dataset when ResNet18 is the desired model}
\label{tb:Office31_result_18}
\centering
\resizebox{\textwidth}{!}{
\begin{tabular}{|l||r|r|r|r|r|r||r|}
\hline
\textbf{Training methods}            & \textbf{A $\xrightarrow{}$ W} & \textbf{W $\xrightarrow{}$ A} & \textbf{D $\xrightarrow{}$ W} & \textbf{W $\xrightarrow{}$ D} & \textbf{D $\xrightarrow{}$ A} & \textbf{A $\xrightarrow{}$ D} & \textbf{Average}     \\ \hline \hline
Baseline 1: UDA $\xrightarrow{}$ KD from ResNet50             & 28.8       & 5.8       & 33.7       & 51.1       & 7.8       & 25.1       & 25.3         \\ \hline
Baseline 2: KD $\xrightarrow{}$ UDA from ResNet50 & 69.0 & 57.3 & 96.2 & \textbf{100} & 56.3 & 73.6 & 75.4 \\ \hline
Baseline 3: UDA only on ResNet18   & 60.2   & 49.2   & 93.7   & 97.7   & 47.6   & 66.4   & 69.1 \\ \hline \hline
TCP: 45\% pruned from ResNet50  & 77.4   & 46.3   & \textbf{96.3}   & \textbf{100}    & 36.1   & 72.0     & 71.3       \\ \hline
Ours: ResNet18 from ResNet50  & 78.9   & 56.8   & 93.8   & \textbf{100}    & 56.0     & \textbf{81.7}   & 77.8 \\ \hline
Ours: ResNet18 from ResNet101 & \textbf{79.2}   & \textbf{58.1}   & 94.2   & \textbf{100}    & \textbf{57.2}   & 79.9   & \textbf{78.1}        \\ \hline
\end{tabular}
}
\end{table*}

In these experiments, the images are cropped to a fixed resolution of 224x224. Regarding the hyper-parameters, we use a starting $\beta$ value of $0.1$ and an end value of $0.9$. For the KD hyper-parameters, we use a temperature $\tau = 20$ and $\alpha = 0.8$. Overall, we use a learning rate starting at $0.001$ for UDA optimizer in Office31 and $0.0001$ for ImageClef-DA, $0.001$ for KD optimizer for both datasets, a momentum of $0.9$ and $400$ epochs.

Our implementation can be found on-line at: \url{https://github.com/LIVIAETS/KD_UDA}

\section{Results and Discussion}

\subsection{Results on Office31:}
For the Office31 dataset, our results outperform most of the baseline and the current existing techniques. From Tables \ref{tb:Office31_result_34} and \ref{tb:Office31_result_18}, we see that, the third baseline "$UDA$ only" performs better than $UDA \xrightarrow{} KD$. This can be explained by the fact that there is no label for target dataset and the distillation loss alone is not sufficient, as there exist a need for supervising the cross-entropy loss. The result of the second baseline is better than the rest of the baselines and TCP, which can be explained because the student model resulting from KD is already trained on a labeled source data distribution. Finally, our techniques using both teacher ResNet50 and ResNet101 perform better than TCP. In both ResNet34 and ResNet18 settings, the difference between the average of our results and TCP is considerable. This shows that combining UDA and KD in a progressive setting brings additional benefits. Between ResNet34 with teacher ResNet50 and ResNet101, there exists a slight difference. We notice that the student model with a larger teacher model performs slightly better. This is expected, since ResNet101 has a better generalization capability than ResNet50. Lastly, our technique also seems to improve UDA since our methods perform better than the second baseline, which only performs UDA on a compact model.

\subsection{Results on ImageClef-DA:} 
For this dataset, the results obtained by our methods are shown in Table \ref{tb:ImageClef_result_34} and Table \ref{tb:ImageClef_result_18}, which demonstrate that the proposed technique outperforms both the baselines and current existing techniques. In contrast, the results obtained with ResNet34 are closer to TCP and the baseline than in the previous dataset, whereas our result with ResNet18 shows a bigger gap between the baselines and the proposed techniques. This can be first explained by the fact that ImageClef-DA is a better balanced dataset where each subset has the same number of samples. Secondly, the third baseline of "UDA only on a ResNet18" already performs better than TCP, which may explain why our methods have a better performance. In this table, the difference between teacher models is closer, this shows that a bigger teacher may not help in improving performance since the learning bottleneck is now on the student model. 

\begin{table*}[h!]
\caption{Accuracy of proposed and baseline methods on ImageClef-DA dataset when ResNet34 is the desired model}
\label{tb:ImageClef_result_34}
\centering
\resizebox{\textwidth}{!}{
\begin{tabular}{|l||r|r|r|r|r|r||r|}
\hline
\textbf{Training methods}            & \textbf{I $\xrightarrow{}$ P} & \textbf{P $\xrightarrow{}$ I} & \textbf{I $\xrightarrow{}$ C} & \textbf{C $\xrightarrow{}$ I} & \textbf{C $\xrightarrow{}$ P} & \textbf{P $\xrightarrow{}$ C} & \textbf{Average} \\ \hline \hline
Baseline 1: UDA $\xrightarrow{}$ KD from ResNet50             & 48.0       & 41.0       & 46.0       & 39.6       & 38.8       &    39.0    & 42.0        \\ \hline
Baseline 2: KD $\xrightarrow{}$ UDA from ResNet50 & 76.6 & 87.3 & 92.0 & 80.0 & 65.6 & 90.3 & 81.9 \\ \hline
Baseline 3: UDA only on ResNet34   & 73.3  & 86.3   & 92.6   & 79.3   & 65.8   & 87.5   &    80.8     \\ \hline \hline
TCP: 12\% pruned from ResNet50 & 75.0  & 82.6   & 92.5   & 80.8   & 66.2   & 86.5   & 80.6    \\ \hline
Ours: ResNet34 from ResNet50  & \textbf{75.6}  & \textbf{89.0}   & 92.6   & \textbf{83.8}   & \textbf{66.5}   & \textbf{92.8}   & \textbf{83.3}    \\ \hline
Ours: ResNet34 from ResNet101 & 75.0  & 87.6   & \textbf{93.3}   & 83.5   & \textbf{66.6}   & 91.8   & \textbf{83.2}    \\ \hline
\end{tabular}
}
\end{table*}

\begin{table*}[h!]
\caption{Accuracy of proposed and baseline methods on ImageClef-DA dataset when ResNet18 is the desired model}
\label{tb:ImageClef_result_18}
\centering
\resizebox{\textwidth}{!}{
\begin{tabular}{|l||r|r|r|r|r|r||r|}
\hline
\textbf{Training methods}            & \textbf{I $\xrightarrow{}$ P} & \textbf{P $\xrightarrow{}$ I} & \textbf{I $\xrightarrow{}$ C} & \textbf{C $\xrightarrow{}$ I} & \textbf{C $\xrightarrow{}$ P} & \textbf{P $\xrightarrow{}$ C} & \textbf{Average}     \\ \hline \hline
Baseline 1: UDA $\xrightarrow{}$ KD from ResNet50             & 45.1       & 41.8    & 42.5       & 43.1       &  43.3  & 34.5 & 41.7             \\ \hline
Baseline 2: KD $\xrightarrow{}$ UDA from ResNet50 & 72.1 & 86.3 & 91.8 & 74.6 & 61.8 & 90.6 & 79.5 \\ \hline
Baseline 3: UDA only on ResNet18   & 70.6   & 83.8   & 86.1   & 75.3   & 62.0     & 89.1   & 77.8 \\ \hline \hline
TCP: 45\% pruned from ResNet50  & 67.8   & 77.5   & 88.6   & 71.6   & 57.7   & 79.5   & 73.7 \\ \hline
Ours: ResNet18 from ResNet50  & 73.1   & 88.0   & 92.1   & \textbf{77.3}   & \textbf{65.6}   & \textbf{91.0}   & \textbf{81.1}     \\ \hline
Ours: ResNet18 from ResNet101 & \textbf{73.5}   & \textbf{88.6}   & \textbf{92.3}   & 76.8   & 64.1   & \textbf{91.1}   & \textbf{81.0}       \\ \hline
\end{tabular}
}
\end{table*}


\subsection{Computational Complexity:} 
Comparison in terms of complexity is depicted in Table \ref{tb:complexity_comp}. While pruned TCP\cite{TCP} models have fewer parameters than our student models, we achieve the same number of FLOPS on ResNet34, and fewer FLOPs on ResNet18. This means that while TCP prunes more parameters, it may not have a lot of impact on the number of FLOPS since the pruned filters are ranked and pruned globally across the network instead of being pruned at each layer. This also means that TCP prunes away filters that do not impact the FLOPS, but that can impact performance, which may hamper the global objective. Another important point of having more parameters is that, over-parametrization can help generalization, increasing the chances of our student model to have a better generalization than a pruned model with less parameters. 

\begin{table}[h!]
\caption{Computational complexity of proposed and TCP networks.}
\label{tb:complexity_comp}
\resizebox{\columnwidth}{!}{
\begin{tabular}{|l||c|c|c|}
\hline
\textbf{Models} & \textbf{no. operations}  & \multicolumn{2}{|c|}{\textbf{no. parameters (M)}}\\
                &   \textbf{(GFLOPS)}  & \textbf{Office31}  & \textbf{ImageClef} \\ \hline \hline
Teacher/Original: ResNet50  & 4.1   & 25.5  & 25.5  \\ \hline
TCP:                        &       &       &   \\ 
- 12\% pruned from ResNet50  & 3.6  & 15.8  & 15.9  \\ 
- 45\% pruned from ResNet50  & 2.2  & 10.6  & 10.9  \\ \hline
Student: from ResNet50      &       &       &    \\  
- ResNet34                  & 3.6   & 21.7  & 21.7  \\  
- ResNet18                  & 1.8   & 11.1  & 11.1  \\ \hline
\end{tabular}
}
\end{table}

\subsection{Comparison over larger teacher and student:}

In this section, we select the scenario of $I \xrightarrow{} C$ and increase the teacher model to ResNet152. Then, we apply our algorithm on this model with several student models going from ResNet18 to Resnet101. We compare the results from these experiments with our previous ones, along with the result of performing UDA using only Eq.\ref{eq:teacher_mmd}.

\begin{figure}[h]
    \centering
    \includegraphics[width=9cm]{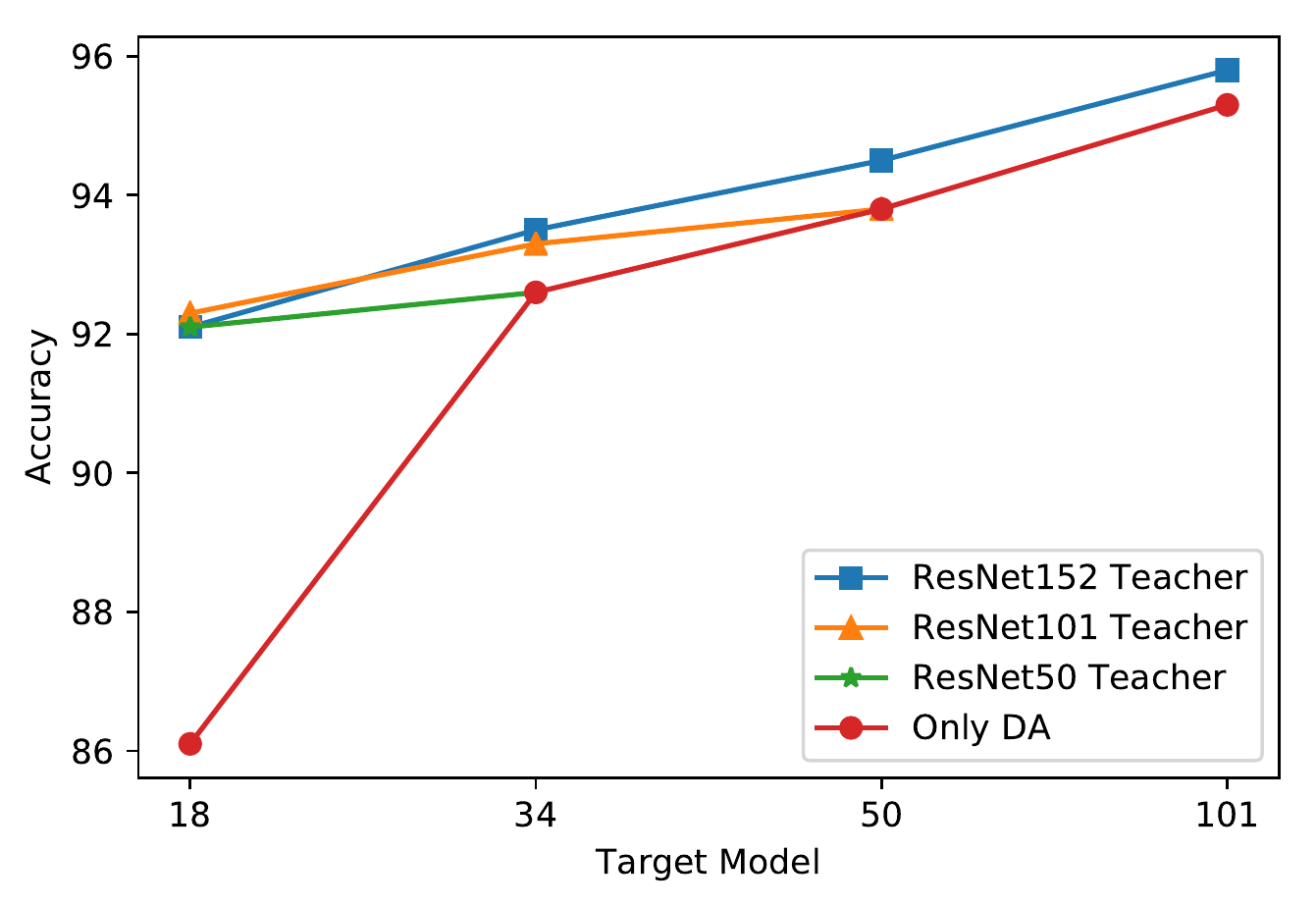}
    \caption{Comparison of different teacher and student models.}
    \label{fig:Resnet152_Teacher}
\end{figure}

From the Fig.\ref{fig:Resnet152_Teacher}, we see that, having a larger teacher only has a slight performance increase compared to the smaller teacher. Furthermore, we also noticed that there is no degradation in terms of performance when the teacher is ResNet152 and the student is ResNet18. Finally, having a ResNet50 as a student model definitely helps increasing the performance, which makes sense since we are using a much larger model. Overall, while having a larger teacher does not seem to have a big impact on accuracy, it can negatively impact the training time and increases the risk of overfitting.

\section{Conclusion}

In this paper, we proposed a combination of KD and UDA that remains unexplored in literature and tackles both the problem of domain shift and model complexity. Our results suggest that the proposed method is capable of obtaining a compressed model adapted to an unsupervised target domain that performs better than state-of-the-art method and current baselines. Additionally, our progressive technique is capable of having a big gap between the student and the teacher without suffering having a performance degradation on the student. The proposed technique is generic and should be able to work with most of current UDA and KD techniques, in future works, we will evaluate our method on other KD and UDA techniques.

\section*{Acknowledgement}

This research was partially supported by the Mathematics of Information Technology and Complex Systems (MITACS) and the Natural Sciences and Engineering Research Council of Canada (NSERC) organizations. 

{
\bibliographystyle{ieeetran}
\bibliography{egbib}
}

\end{document}